\documentclass[journal]{IEEEtran}
\usepackage{amsmath,amsfonts}
\usepackage{algorithmic}
\usepackage{algorithm}
\usepackage{array}
\usepackage[caption=false,font=normalsize,labelfont=sf,textfont=sf]{subfig}
\usepackage{textcomp}
\usepackage{stfloats}
\usepackage{url}
\usepackage{verbatim}
\usepackage{rotating}
\usepackage{graphicx}
\usepackage{cite}
\usepackage{tablefootnote}
\usepackage{scalerel}
\usepackage{multicol}
\usepackage{multirow}
\usepackage{tikz}

\usepackage{xcolor}
\usepackage{soul}

\usetikzlibrary{svg.path}
\DeclareMathOperator*{\argmin}{arg\,min}
\definecolor{orcidlogocol}{HTML}{A6CE39}
\tikzset{
  orcidlogo/.pic={
    \fill[orcidlogocol] svg{M256,128c0,70.7-57.3,128-128,128C57.3,256,0,198.7,0,128C0,57.3,57.3,0,128,0C198.7,0,256,57.3,256,128z};
    \fill[white] svg{M86.3,186.2H70.9V79.1h15.4v48.4V186.2z}
                 svg{M108.9,79.1h41.6c39.6,0,57,28.3,57,53.6c0,27.5-21.5,53.6-56.8,53.6h-41.8V79.1z M124.3,172.4h24.5c34.9,0,42.9-26.5,42.9-39.7c0-21.5-13.7-39.7-43.7-39.7h-23.7V172.4z}
                 svg{M88.7,56.8c0,5.5-4.5,10.1-10.1,10.1c-5.6,0-10.1-4.6-10.1-10.1c0-5.6,4.5-10.1,10.1-10.1C84.2,46.7,88.7,51.3,88.7,56.8z};
  }
}

\newcommand\orcidicon[1]{\href{https://orcid.org/#1}{\mbox{\scalerel*{
\begin{tikzpicture}[yscale=-1,transform shape]
\pic{orcidlogo};
\end{tikzpicture}
}{|}}}}

\usepackage{bbding}
\usepackage{pifont}
\usepackage{wasysym}
\usepackage{amssymb}

\usepackage[unicode=true,
 bookmarks=true,bookmarksnumbered=true,bookmarksopen=true,bookmarksopenlevel=1,
 breaklinks=false,pdfborder={0 0 0},pdfborderstyle={},backref=false,colorlinks=true,
            linkcolor = black,
            urlcolor  = black,
            citecolor = black,
            anchorcolor = black] {hyperref}
            
\colorlet{SB}{purple!50!black}   
\colorlet{DS}{blue!50!black}     
\colorlet{FB}{green!50!black}    

\begin{document}

\title{DH-PTAM: A Deep Hybrid Stereo Events-Frames Parallel Tracking And Mapping System}

\author{Abanob Soliman$\,^{\orcidicon{0000-0003-4956-8580}}$, Fabien Bonardi$\,^{\orcidicon{0000-0002-3555-7306}}$, D\'esir\'e Sidib\'e$\,^{\orcidicon{0000-0002-5843-7139}}$, and Samia Bouchafa$\,^{\orcidicon{0000-0002-2860-8128}}$}



\maketitle

\begin{abstract}
This paper presents a robust approach for a visual parallel tracking and mapping (PTAM) system that excels in challenging environments. Our proposed method combines the strengths of heterogeneous multi-modal visual sensors, including stereo event-based and frame-based sensors, in a unified reference frame through a novel spatio-temporal synchronization of stereo visual frames and stereo event streams. We employ deep learning-based feature extraction and description for estimation to enhance robustness further. We also introduce an end-to-end parallel tracking and mapping optimization layer complemented by a simple loop-closure algorithm for efficient SLAM behavior. Through comprehensive experiments on both small-scale and large-scale real-world sequences of VECtor and TUM-VIE benchmarks, our proposed method (DH-PTAM) demonstrates superior performance in terms of robustness and accuracy in adverse conditions, especially in large-scale HDR scenarios. Our implementation's research-based Python API is publicly available on GitHub for further research and development: \url{https://github.com/AbanobSoliman/DH-PTAM}.
\end{abstract}


\section{Introduction}

\IEEEPARstart{S}{imultaneous} Localization and Mapping (SLAM) is pivotal in robotics and computer vision, aiming to chart unknown terrains while discerning an agent's location. Among the key SLAM contributions is the Parallel Tracking and Mapping (PTAM) method \cite{4538852}, which uniquely separates tracking and mapping into parallel threads for augmented efficiency and real-time performance in monocular systems. However, PTAM faced scale ambiguity challenges inherent to monocular SLAM. Addressing this issue, its successor, Stereo Parallel Tracking and Mapping (S-PTAM) \cite{pire2017s}, leveraged stereo vision to extract depth information directly, eliminating scale ambiguity and fortifying robustness. In recent years, learning-based features extraction and description methods \cite{detone2018superpoint,revaud2019r2d2}, and deep learning based approaches \cite{teed2021droid} have been applied to improve robustness.

\begin{figure}[!t]
\centering
\includegraphics[width=\linewidth]{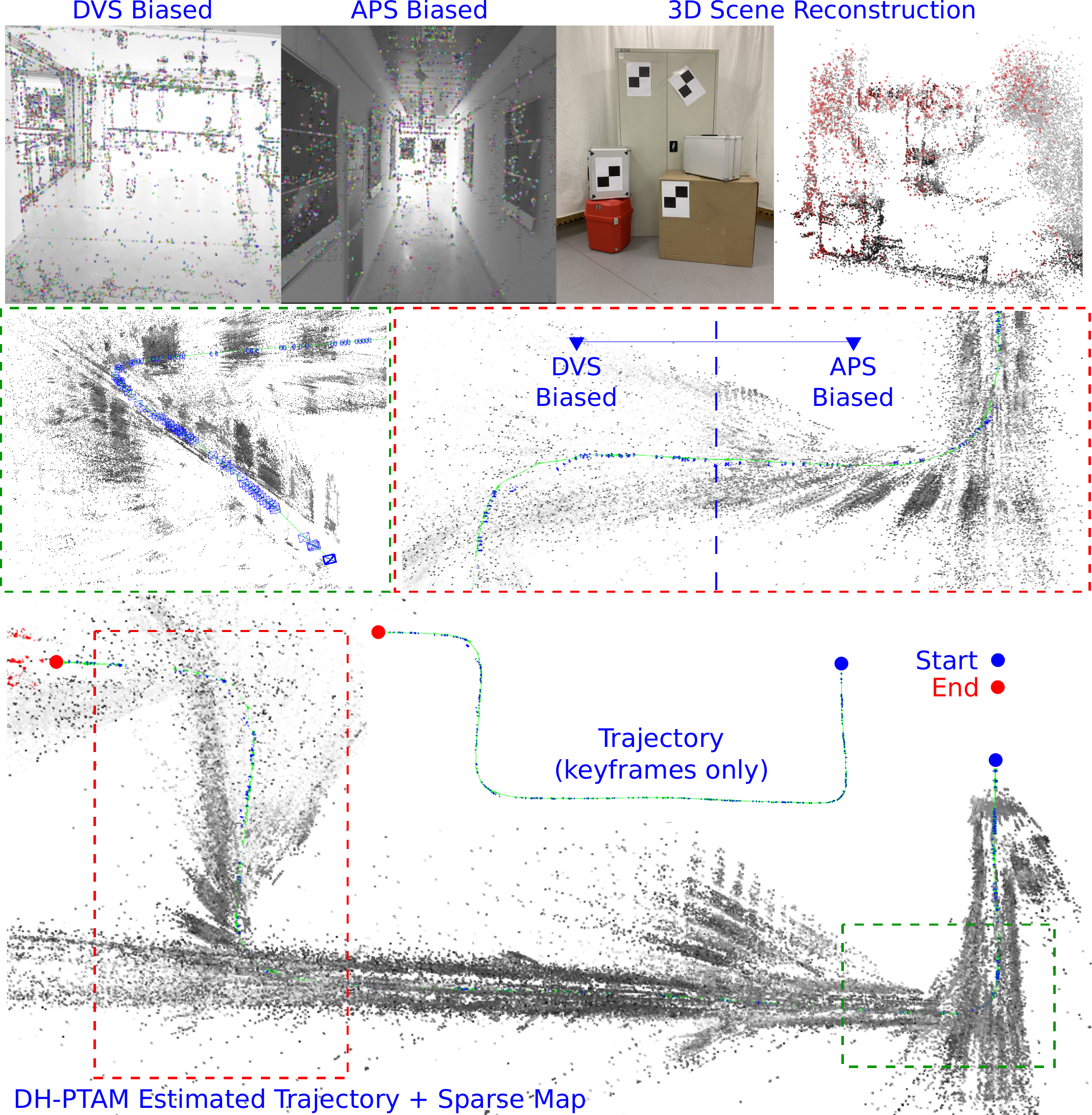}
\caption{Experiments on school-scooter and corner-slow sequences from the VECtor dataset show the estimated trajectory with the constructed scene map (green dotted rectangle). The red dotted rectangle highlights an HDR use-case where DH-PTAM estimates the trajectory continuously based on the two fusion modes (Dynamic Vision Sensor (DVS) or Active Pixel Sensor (APS) biased). APS: denotes the standard camera's global shutter frames.}
\label{fig:teaser}
\end{figure}

Visual Odometry (VO), an integral component of SLAM, has predominantly depended on conventional cameras to determine motion. However, such methods often fail in high dynamic range (HDR) scenarios where lighting conditions fluctuate \cite{zhang2017active} (see Fig. \ref{fig:teaser}). Fortunately, the innovation of event cameras, also termed asynchronous or dynamic vision sensors (DVS) \cite{hidalgo2022event}, offers a groundbreaking solution. Unlike their traditional counterparts that capture frames at fixed intervals, event cameras relay a continuous stream of "events" showcasing pixel-wise brightness alterations. This not only empowers them to operate at unparalleled speeds and in dimly lit conditions but also markedly diminishes motion blur \cite{sun2021autonomous}. 

With the ability to adeptly handle swift motions, scenes with significant HDR, and challenging lighting, event cameras emerge as superior alternatives, increasing both SLAM and VO's robustness and precision in environments that would confound traditional cameras. The distinctive event-based design makes these sensors incredibly adapted at tracking fast-moving objects, setting them apart as an invaluable asset to visual odometry, especially under adverse conditions.

Deep learning-based features are more robust than traditional methods \cite{detone2018superpoint,revaud2019r2d2}, as they can learn from large amounts of data and generalize well to unseen data. They are also more invariant to changes in viewpoint and lighting, making them suitable for real-world applications. Recently, pre-trained models have been widely adopted in computer vision and have achieved state-of-the-art performance in object detection, semantic segmentation, and image classification tasks.

This paper proposes a deep hybrid stereo events-frames parallel tracking and mapping system  (see Fig. \ref{fig:pipeline}) that significantly improves SLAM accuracy and robustness in dynamic environments. This system combines the advantages of stereo RGB and event cameras, which can capture visual information at high temporal resolution. The use of deep learning techniques in this system allows for the extraction of robust features from the stereo hybrid image and event frames, which improves the accuracy of the feature-matching process and the estimation of the camera pose. 

The main contributions can be summarized as follows: 

\begin{itemize}
    \item We propose an end-to-end parallel tracking and mapping (PTAM) approach based on a novel spatio-temporal synchronization of stereo visual frames with event streams.
    \item We propose a simple mid-level feature loop-closure algorithm for prompt SLAM behavior based on a learning-based feature description method to maximize robustness.
    \item DH-PTAM's effectiveness is evaluated in both stereo event-aided and image-based visual SLAM modes, achieving improved accuracy when incorporating event information, shown in an ablation study on the CPU versus the GPU of a consumer-grade laptop.
\end{itemize}

This paper is organized as follows: Section \ref{sec:SOTA} gives a brief overview of the state-of-the-art SLAM methods. Section \ref{sec:method} provides an in-detail overview of the proposed method and offers insights into the novel parts of the algorithm. Section \ref{sec:eval} comprehensively evaluates the algorithm on the most recent VECtor \cite{gao2022vector} and TUM-VIE \cite{klenk2021tum} benchmarks, along with defining the limitations. Section \ref{sec:concl} summarizes the experiments' main observations, the proposed method's behavioral aspects, and the start points for future works.

\section{Related Work\label{sec:SOTA}}

\begin{figure*}[!t]
\centering
\includegraphics[width=\linewidth]{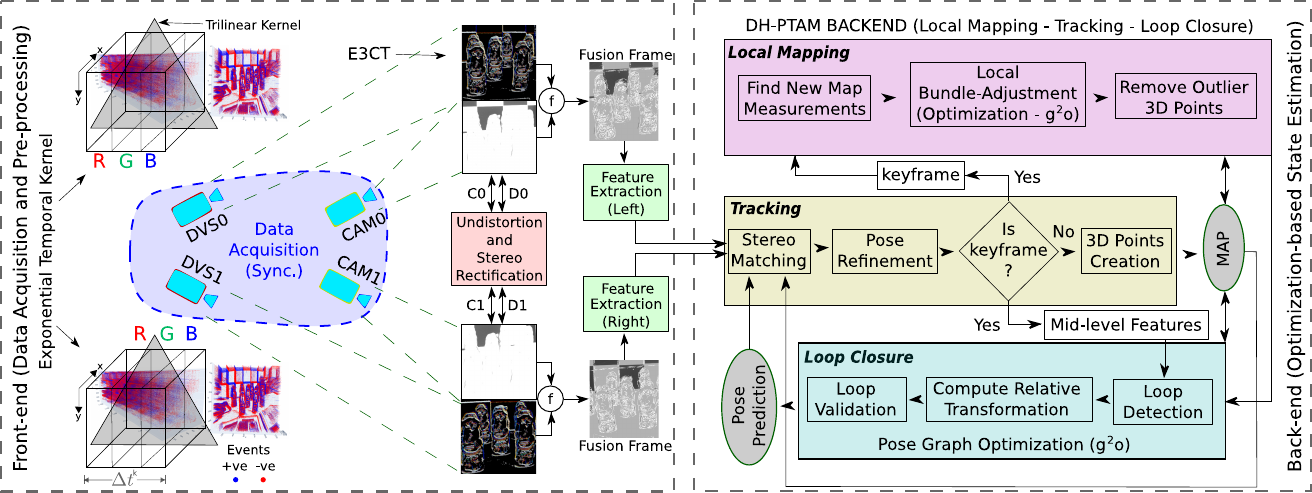}
\caption{Block diagram of the proposed hybrid event-aided stereo visual odometry approach (DH-PTAM). $f$ denotes the fusion function defined in \eqref{eqn:fusion}. $\Delta{t}^k$ is the event volume $\mathcal{V}_0(x,y,t)$ accumulation time defined in \eqref{eqn:volume}. $\text{E3CT}$ denotes the Event 3-Channel Tensor \cite{ibiscape}, an image-like event representation.}
\label{fig:pipeline}
\end{figure*}

\textbf{Events-Frames hybridization}. Event-aided systems leverage the high-quality representations that events can produce after processing, especially in dynamic and dimmed environments where RGB camera frames fail. Some of the well-known event representations are event image (EI) \cite{rebecq2016evo}, Time Surfaces (TS) \cite{sironi2018hats}, Event Spike Tensor (EST) \cite{gehrig2019end}, and recently Event 3-Channel Tensor (E3CT) \cite{ibiscape}. Others \cite{hidalgo2022event} build the front-end on an Event Generation Model (EGM) \cite{gehrig2020eklt} or construct motion-compensated event frames (MEF) \cite{vidal2018ultimate} aided by a gyroscope. Towards a traditional frame reconstruction from events, \cite{kim2016real} proposes a Log Intensity Reconstruction (LIR), a model-based method, and \cite{cadena2021spade} proposes Spade-e2vid, a learning-based method.

Indirect methods \cite{campos2021orb}, such as frame-based approaches, extract keypoints from the input data in the front-end. This front-end stage typically involves detecting and matching salient features in the sensory data, such as images or event streams. These keypoints are then passed to the back-end, where state estimation algorithms are used to estimate the robot's pose and build a consistent map of the environment. 

Conversely, direct methods \cite{hidalgo2022event} attempt to process all available sensor data, such as individual pixel intensity changes in images (events) or all RGB frame pixels, without any intermediate filtering or feature extraction in the front-end, relying on the back-end to handle the entire data. The proposed method adopts a hybrid approach where all events are directly processed during the events-frames fusion in the front-end, while only the reliable learning-based features from the fusion frames are fed to the back-end (see Fig. \ref{fig:pipeline}).

Table \ref{tab:SOTA} compares the latest event-based and event-aided VO solutions concerning the sensor setup, events pre-processing layer (EPL), direct or indirect event processing, and the loop-closure capability to minimize visual drifts.

\textbf{Event-aided visual-SLAM}. DH-PTAM builds upon the pioneering work of \cite{hidalgo2022event}, which introduces a monocular 6-DoF visual odometry model that synergistically integrates events and grayscale frames using a direct approach. In this model, the fusion of events and frames is achieved by applying EGM to the intensity frame, transforming it into a unified reference that aligns with the event stream, termed as brightness increments (event-like). This process is followed by an error minimization cost function, which estimates the extrinsics in real-time for accurate mapping and is augmented by a photometric bundle adjustment for tracking. However, DH-PTAM distinguishes itself from \cite{hidalgo2022event} by employing a image-like unified reference in its front-end, harmoniously combining stereo event streams with stereo RGB frames.

Loop-closure detection is paramount in visual-SLAM, effectively minimizing drifts by allowing a system to recognize previously traversed locations. The realm of loop-closure detection predominantly features two methodologies: mid-level features \cite{koniusz2013comparison} and the bag-of-words model \cite{GalvezTRO12}. While mid-level features offer a more nuanced representation than low-level features such as edges and corners, they don't consider the specificity of high-level features like object recognition. Deep learning descriptors, as referenced in \cite{app10010140}, typify mid-level features. They extract more sophisticated information from raw data than low-level features like pixel values, yet they don't reach the specificity of direct task-related features, for instance, object labels.

\begin{table}[!t]
\caption{Direct and Indirect (D/I) Visual Odometry methods based (B) on events and/or aided (A) by events \label{tab:SOTA}}
\begin{center}
\resizebox{\linewidth}{!}{
\begin{tabular}{lccccl}
\hline
Method & B/A & D/I & EPL$^a$ & LC$^b$ & More Information\\
\hline
Kim \cite{kim2016real} & B & D & LIR & $\times$ & 3 EKFs + Image reconst.\\
Rebecq \cite{rebecq2016evo} & B & D & EI & $\times$ & Monocular (PTAM)\\
Zhou \cite{zhou2021event} & B & D & TS & $\times$ & Stereo (PTAM)\\
Kueng \cite{kueng2016low} & A & I & $\times$ & $\times$ & Event-aided Tracking\\
Rosinol \cite{vidal2018ultimate} & A & I & MEF & $\times$ & Mono + IMU (front-end)\\
Hidalgo-Carri\'o \cite{hidalgo2022event} & A & D & EGM & $\times$ & Monocular Odometry\\
\hline
\textbf{Proposed} & A & D/I & E3CT & \checkmark & Stereo (PTAM) + DL$^c$\\
\hline
\end{tabular}
}
\end{center}
\footnotesize{$^a$ denotes an Event Pre-processing Layer. $^b$ denotes Loop-Closure capability. $^c$ denotes the only method incorporating Deep Learning-aided features.}
\end{table}

\section{Methodology\label{sec:method}}

\subsection{System Overview}

Fig. \ref{fig:pipeline} illustrates the main components and the process of DH-PTAM. The system establishes a global reference frame based on the camera position in the initial frame. A preliminary map is created by identifying and triangulating distinctive points in the first stereo pair of images. For subsequent frames, the tracking thread calculates the 6D pose of each stereo frame by minimizing the discrepancy between the projected map points and their matches. The system chooses a subset of keyframes used in another thread to update the map at a slower pace. 

Map points are derived from the stereo matches of each keyframe and added to the map. The mapping thread constantly improves the local discrepancy by adjusting all map points, and stereo poses using Bundle Adjustment. A pose graph is utilized to preserve the global consistency of the map which is a shared resource among the tracking, mapping, and loop-closing threads. Point correspondences are actively searched between keyframes to strengthen the constraints of the pose graph optimization smoothing process.

\textbf{Notations}. The odometry state representation comprises the 3D points $X_w^k$ and a 7-increment vector $\mu \in \mathfrak{se}(3)$, which is the current pose of the left fusion frame at time $k$:

\begin{equation}
\mu^k=\left[ \delta x \; \delta y \; \delta z \; \delta q_{x} \; \delta q_{y} \; \delta q_{z} \; \delta q_{w} \right]^\top\;,
\end{equation} 

\noindent where $[\delta x \; \delta y \; \delta z]^\top$ is the incremental translation vector and $[\delta q_{x} \; \delta q_{y} \; \delta q_{z} \; \delta q_{w}]^\top$ is the incremental quaternion vector. 

\subsection{Temporal Synchronization Approach}

\begin{figure}[!t]
\centering
\includegraphics[width=\linewidth]{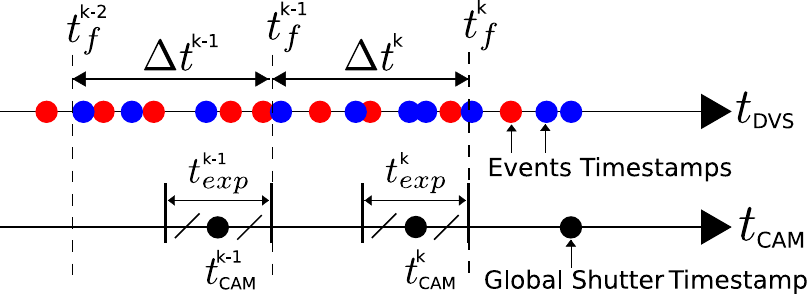}
\caption{Temporal synchronization scheme. $t_{exp}$ is the global shutter camera exposure time. $\Delta{t}$ is the event representation (E3CT) volume accumulation window. $t_f$ is the fusion frame calculated timestamp. $t_{\text{DVS},\text{CAM}}$ are the DVS events, and RGB camera frames timestamps, respectively.}
\label{fig:sync}
\end{figure}

\begin{figure}[!b]
\centering
\includegraphics[width=\linewidth]{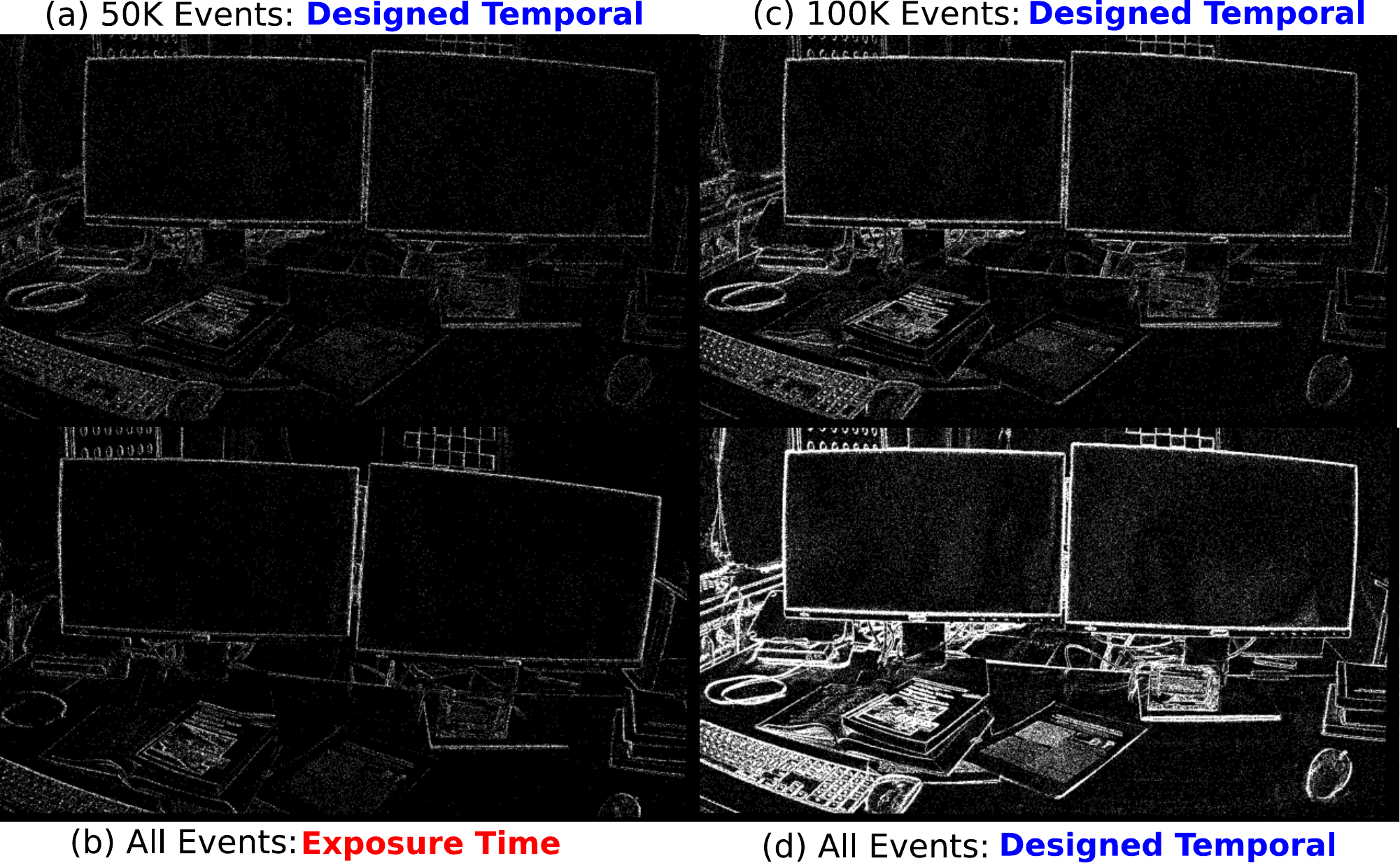}
\caption{Ablation study on reducing the temporal window width versus controlling the number of events in the designed window. All event frames are post-processed E3CTs by median filtering followed by a binary threshold.}
\label{fig:samples}
\end{figure}

Our temporal synchronization approach (see Fig. \ref{fig:sync}) considers the general case of global shutter cameras where the exposure time $t_{exp}$ is known. We adopt the constant-time $\Delta{t}^k$ events accumulation window $k$ approach where the number of accumulated events during this temporal window is ablated in the qualitative analysis in Fig. \ref{fig:samples}. 

As soon as stereo RGB camera frames are received at timestamps $t_{\text{CAM}}$, we calculate the fusion frames timestamps assuming the hardware synchronization of stereo RGB images and stereo event streams, using:

\begin{equation}\label{eqn:volume}
    t_{f}=t_{\text{CAM}}+\frac{t_{exp}}{2}\;,\;\Delta{t}^k=t_{f}^k-t_{f}^{k-1}\;, 
\end{equation}

\noindent where the left $t_{\text{CAM}}$ is the selected stereo keyframe timestamp.

\subsection{Spatial Hybridization Approach}

Leveraging the findings of \cite{9710962}, where Scale-Consistent monocular depth learning is explored, our method is firmly rooted in the depth scale equivalence principle. Rather than just an assumption, this principle is an inference drawn from the inherent properties of closely-spaced APS-DVS sensors. It asserts that the depth scales between frames maintain a consistent relationship. Diverging from the approach in \cite{9710962}, we incorporate the front-end's depth uncertainty into the back-end's stereo bundle adjustment process. Here, it acts as an adaptive mechanism, progressively refining depth estimates, thereby boosting system performance. The E3CT events pre-processing layer is adopted and modeled as two consecutive filtering kernel convolutions on the event volume $\mathcal{V}_0(x,y,t)$ of temporal width $\Delta{t^k}$ (see Fig. \ref{fig:pipeline}). The first kernel to filter the time decaying events in the volume, is the $\alpha$-exponential time decay kernel and is modeled as:

\begin{equation}
    \mathcal{V}_1(x,y,t) \doteq \exp\left(-\alpha\left(\frac{\mathcal{V}_0(x,y,t)-\eta/2}{\eta/6}\right)^2\right) \;,
\end{equation}

\noindent where $\alpha=0.5$ and the decay rate $\eta=30$ [ms] for our model. Followed by a trilinear voting kernel to stack the events in the three channels tensor so that each event contributes to two consecutive channels depending on their location from a vertex of this trilinear kernel. An event near the top contributes a higher weight to the current channel and a lower weight to the neighboring ones. These contribution weights of the three channels can represent a percentage of an R-G-B color map; hence, the E3CT can be considered a synthetic RGB frame of events. The trilinear voting kernel can be modeled as follows:

\begin{equation}
    \mathcal{V}_2(x,y,t_i) \doteq \max\left(0,\,1-\left|\frac{\mathcal{V}_1(x,y,t_i)}{\delta{t}}\right|\right) \;,
\end{equation}

\noindent where $\delta{t}$ is the temporal bin $i$ size as discussed in \cite{gehrig2019end}.

After applying the trilinear temporal voting kernel on the exponential-decay time surface, we stack the 3-channel tensor temporal bins together, resulting in a synthetically colored 2D frame called the Event 3-Channel Tensor (E3CT). In Fig. \ref{fig:pipeline}, we can observe that the constructed synthetic colors are always consistent, meaning that the stereo left and right constructed E3CTs have identical colors for the same scene. 

\begin{figure}[!t]
\centering
\includegraphics[width=\linewidth]{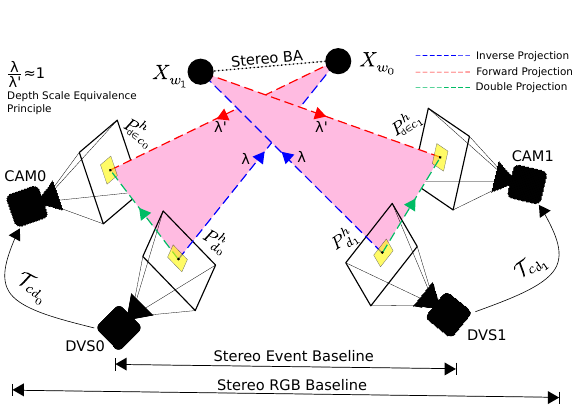}
\caption{Geometry of the stereo hybrid event-RGB cameras stack. $\mathcal{T}_{cd}$ denotes the rigid-body transformations. $P^h_{d\in c}\;\text{and}\;P^h_{d}$ denote pixels locations.}
\label{fig:method-frame}
\end{figure}

Conventional frame-based post-processing operations can be applied to the constructed E3CTs, such as adaptive threshold, contrast stretching, color correction and balance, and denoising functions. We consider a fully calibrated stereo RGB and event cameras stack as represented in Fig. \ref{fig:method-frame}, so that the rigid-body transformations $\mathcal{T}_{cd}$ and the cameras intrinsic parameters $\mathcal{K}_{c},\mathcal{K}_{d}$ are known. 

Given that the same post-processing operations are applied on the current stereo E3CT frames (see Fig. \ref{fig:samples}), the 2D-to-3D-to-2D consecutive inverse-forward projections of the pixels on the E3CT frames $P^h_{d}$ to the RGB camera frames $P^h_{d\in c}$ can be performed as follows (comparable to Equation (2) in \cite{9710962}):

\begin{equation}
    P^h_{d\in c} \doteq \frac{1}{\lambda{'}} \, \mathcal{K}_{c} \, [R|t]_{cd} \, \lambda \, \mathcal{K}_{d}^{-1} \, P^h_{d} + \delta{P^h_{align}} \;,
\end{equation}

\noindent where $(.)^h$ denotes the pixel location in homogeneous coordinates. The term $\delta{P^h_{align}}$ denotes the pixel location alignment correction factor for the RGB and event frames so that the same 3D world point $X^h_{w}$ should correspond exactly to the pixel locations $P^h_{d\in c},\, P^h_{d}$. This alignment term is observed to be constant for the same sensor rig with non-varying intrinsic and extrinsic parameters. $\delta{P^h_{align}}$ value can be estimated using an offline optimization process only once on a selected number of frames (the more the accurate) with high confidence feature matches, and this value is given in Section \ref{sec:eval} for both VECtor and TUM-VIE sequences.

\begin{figure}[!t]
\centering
\includegraphics[width=\linewidth]{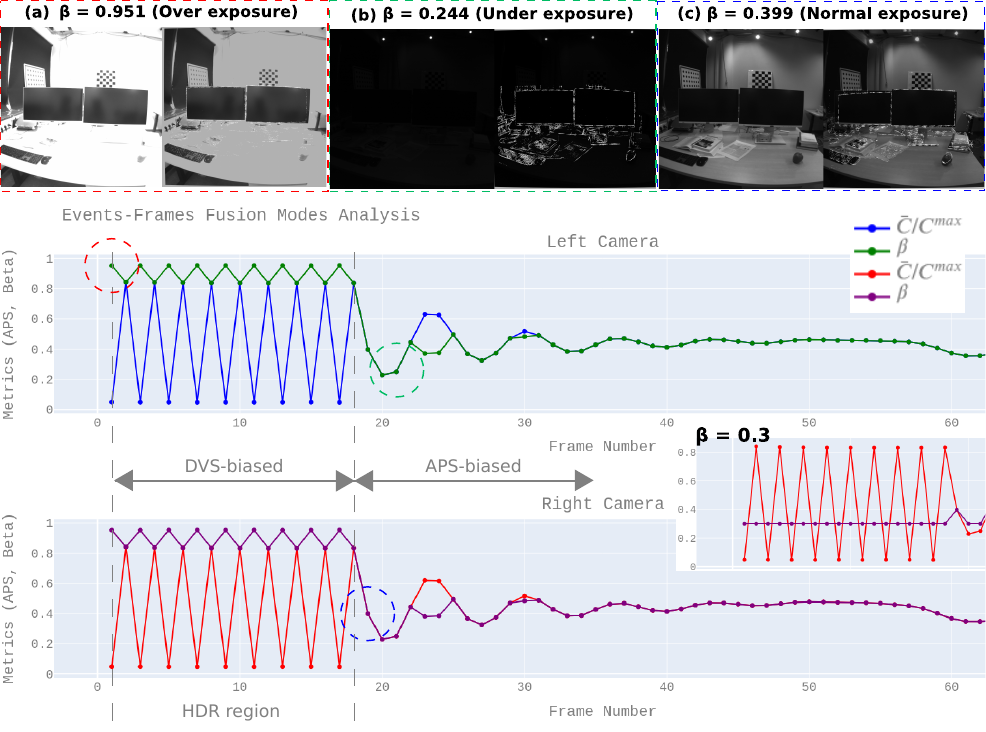}
\caption{Experiments on mocap-desk2 TUM-VIE dataset that show the capability of the proposed events-frames fusion method to maintain and track features in bight and dimmed scenes where grayscale-only frames may fail. For each scenario in top: grayscale frame (left) and fusion frame (right).}
\label{fig:method-circular}
\end{figure}

Finally, the fusion function (and frame) $f(.)$ performs a temporal cross-dissolve (linear blending) between both the left ($D_0, C_0$) and right ($D_1, C_1$) E3CTs and RGB camera frames, respectively, and is formulated as:

\begin{equation}\label{eqn:fusion}
    f(C,D) \,=\, (1-\beta)*C + \beta*D \;,
\end{equation}

\noindent where $\beta\in[0,1]$ is the E3CT contribution weight in the current fusion frame. $\beta$ value is calculated online and depends on the scene lighting and texture conditions. It is set to chirp-shaped values $\beta=\max(\bar{C}/{C^{\text{max}}},\,1-\bar{C}/{C^{\text{max}}})$ when the RGB camera frame fails to detect features due to adverse HDR conditions and low-textured scenes, and this is the DVS-biased fusion mode. For situations where RGB camera frames can detect reliable scene features with good lighting and enough texture, the $\beta$ value is harmonic with the scene lighting conditions according to $\beta=\min(\bar{C}/{C^{\text{max}}},\,1-\bar{C}/{C^{\text{max}}})$, and this is the APS-biased fusion mode. To reduce the amount of extracted features and maintain the back-end processing complexity and latency in reasonable ranges, $\beta$ value can be capped at a certain value, as shown in Fig. \ref{fig:method-circular} by setting $\beta=0.3$ as an example.

 Dynamic scenes with challenging and adverse conditions can easily trigger rapid switching between these two fusion modes during long-term navigation. This causes a critical problem during the feature tracking process using conventional low-level feature detectors, such as ORB, SIFT, SURF, BRIEF, and FAST. Accordingly, applying mid-level feature detectors that depend mainly on learning-based architectures could solve this fusion frame modes alternation problem. We employ the learning-based feature extractors and descriptors \cite{detone2018superpoint,revaud2019r2d2} for their high robustness and feature detection speed. Fig. \ref{fig:method-circular} elaborates in a real-world case-study, the fusion mode automatic switch using the  $\beta$ parameter in response to the rapid intensity fluctuations of the RGB camera frame $C$ in HDR scenarios represented by the parameter $\bar{C}/C^{\text{max}}$.

\subsection{Optimization-based State Estimation}
 
As our work is based on the original S-PTAM system, all the optimization Jacobians mentioned in this section can be found with detailed proofs in \cite{pire2017s}. All objective functions are minimized with the Levenberg-Marquardt algorithm implemented in the $g^2o$ optimization library. We employ the Huber loss function for outliers rejection $\rho(.)$.

\textbf{System bootstrapping}. The first stereo fusion frames are considered a keyframe. Then, a triangulation for the collected feature matches on the left and right fusion frames is performed to initialize the map.

\begin{figure}[!t]
\centering
\includegraphics[width=\linewidth]{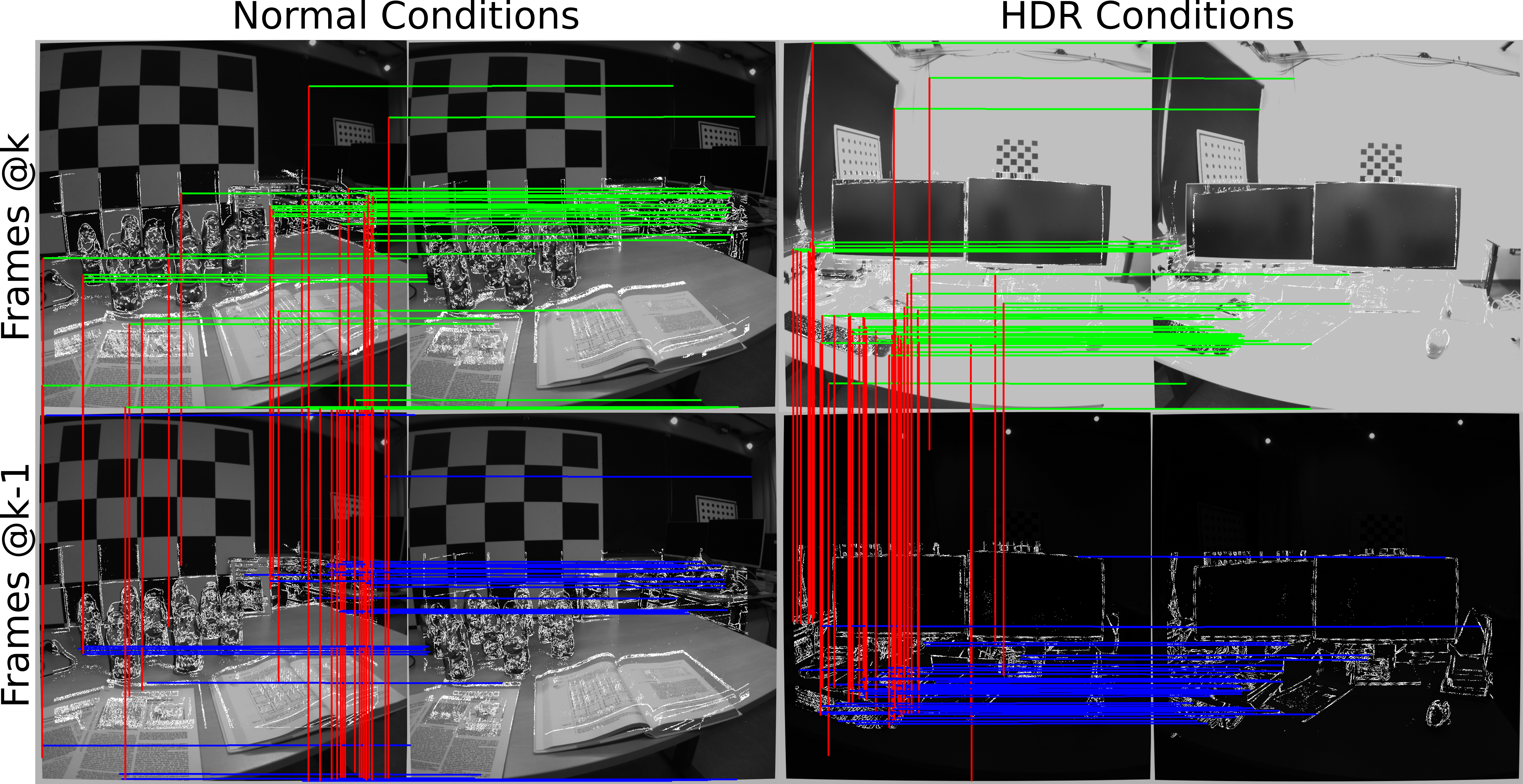}
\caption{Normal versus HDR semi-circular spatio-temporal matching for two stereo consecutive fusion frames. Green and blue lines denote the spatial stereo SuperPoints matching at frames $k$ and $k-1$, respectively. Red lines denote the temporal matching for keypoints of two consecutive keyframes (left).}
\label{fig:method-sync}
\end{figure}

\textbf{Pose tracking thread}. Each map point is projected into the viewing frustum of the anticipated stereo position, and we then look nearby for the match. A valid prediction of the current pose is required for such a projection. By contrasting the descriptions, map points, and features are matched. The $L_2$ norm is computed using the binary descriptors of SuperPoint and R2D2 (see Fig. \ref{fig:method-sync}). The match is valid if the distance falls below a certain threshold; otherwise, it is ignored. The pose refinement is then applied to recover the current pose knowing the previous one using the following objective function:

\begin{equation}
    L^{\text{refine}} = \argmin_{\mu} \sum_{i\in N}\rho(\lvert\lvert 
    J_i^k \mu_k - \Delta{z}_i(\mu_{k-1}, X_w^i)
    \rvert\rvert^2) \;,
\end{equation}

\noindent where $N=\{z_1 , \, \dots \, , \, z_M\}$ and $M$ is the number of matched measurements. The measurement $z=[u,v]^\top$ is a pixel 2D location of the forward projection of a 3D map point $X_w$ using the pinhole model projection function $\pi(X_w^i) = \mathcal{K}_{c} \mathcal{T}^{i}_{f_0 w} X_w^i$. $J_i^k=\frac{\partial \Delta{z}_i(\mu)}{\partial \mu_k}$ is the re-projection error's Jacobian with respect to the current odometry state vector. $\Delta{z}$ is the re-projection error of a matched set of measurements on the current $k$ stereo fusion frames and is defined as:

\begin{equation}
    \Delta{z}_i(\mu, X_w) = z_i - \pi(\exp{(\mu)} \mathcal{T}^{k-1}_{f_0 w} X_w^i) \;,
\end{equation}

\noindent where the 3D point cloud $X_w$ is considered a constant optimization parameter and not updated in the tracking thread and $ \mathcal{T}^{k-1}_{f_0w}=exp(\mu) \in SE(3)$ with $\exp{(.)}$ the exponential map in the $\mathfrak{Lie}$ group for the previous increment state vactor. If the number of observed points is less than 90\% of the points recorded in the previous keyframe, a frame is chosen to be a keyframe after the current pose has been evaluated. Then, new map points are created by triangulating the stereo pair's remaining mismatched features. The keyframe is then placed in the local mapping thread for processing. 

\textbf{Mapping thread}. We apply Bundle Adjustment (BA) to fine-tune the camera poses (keyframe map) and the 3D points (point cloud map). Local Bundle Adjustment minimizes the re-projection error of every point in every keyframe $f^k_{0}$. Given an initial set of $N$ keyframe poses $\{\mathcal{T}_{f_0w}^1, \, \dots \, , \, \mathcal{T}_{f_0w}^N\}$, an initial set of $M$ 3D points $X_w^i$, and measurement sets $S\in \{S_1, \, \dots \, , \, S_N\}$, where each set comprises the measurement $z_i^k$ of the $i^{th}$ point in the $k^{th}$ keyframe, the local BA is performed using the following objective function on all keyframes in a pre-defined sliding-window size $N$: 

\begin{equation}
    L^{\text{BA}} = \argmin_{\mu,\,X_w} \sum_{k=1}^N \sum_{i\in S_k} \rho(\lvert\lvert 
    J_i^k \begin{bmatrix} \mu_k \\ X_w^i \end{bmatrix} - \Delta{z}_i(\mu_{k}, X_w^i)
    \rvert\rvert^2) \;,
\end{equation}

\noindent where the 3D point cloud $X_w$ is considered a variable optimization parameter and is updated in the mapping thread. Hence, the $J_i^k= [ \frac{\partial \Delta{z}_i(\mu_k, X_w^i)}{\partial \mu_k},\, \frac{\partial \Delta{z}_i(\mu_k, X_w^i)}{\partial X_w^i} ]$ is the re-projection error's Jacobian with respect to the current odometry state vector and the 3D point as well.

\textbf{Loop-closure thread}. Instead of the conventional way of keyframe embedding assignments using a bag-of-words, we adopt a simple loop-closure detection method based on the mean of the mid-level learning-based feature descriptors (SuperPoint and R2D2) for each keyframe and assign this mean value as the embedding identity of each keyframe. Once a potential loop closure is detected, the system performs geometric verification through RANSAC-based pose estimation to validate the candidate. If the verification is successful, a loop closure constraint is added to the pose graph, and a graph optimization is performed to distribute the error and update the global map, thus correcting the accumulated drift.

\section{Evaluation\label{sec:eval}}

\begin{figure*}[!htbp]
\centering
\includegraphics[width=\linewidth]{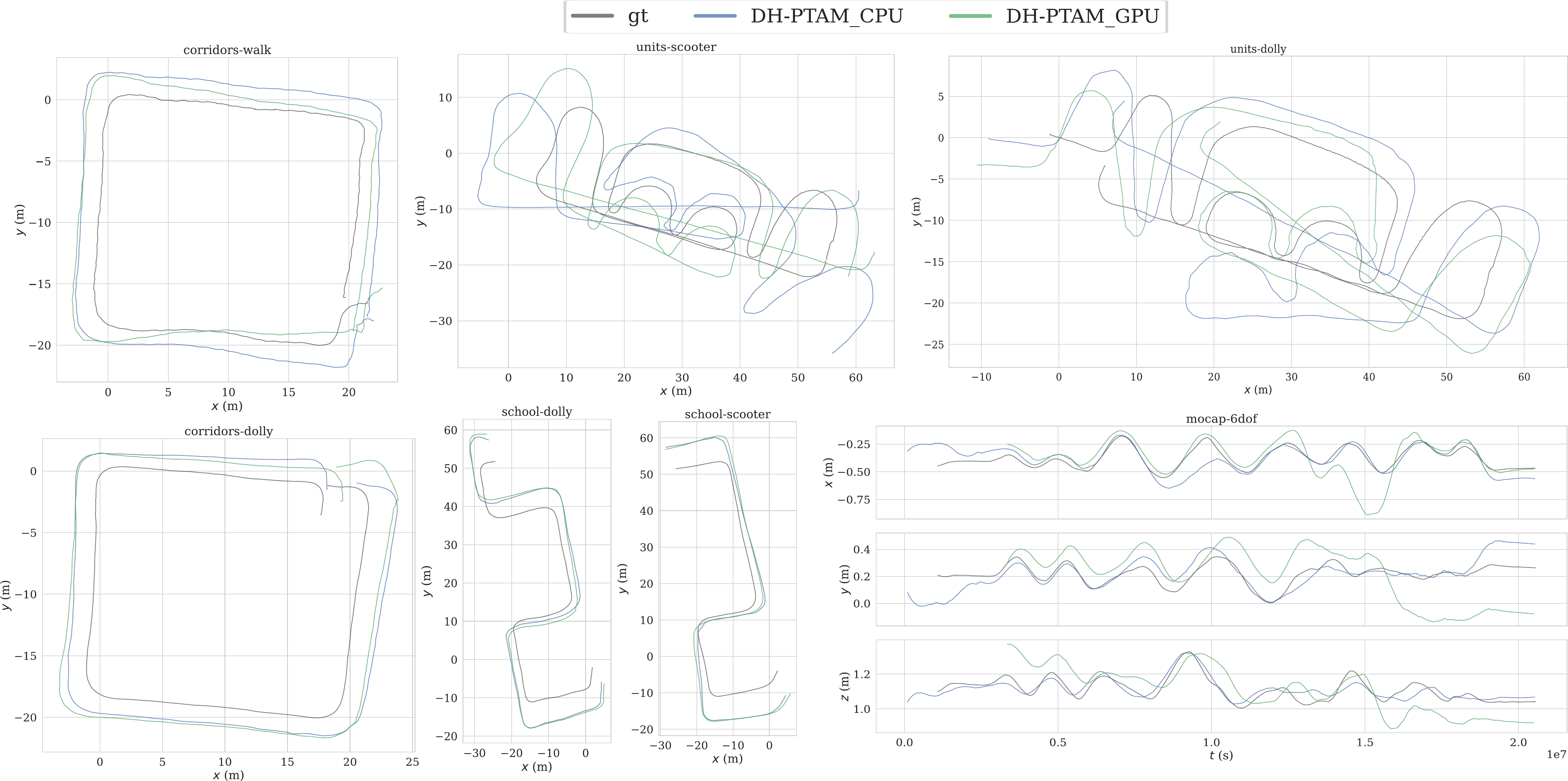}
\caption{DH-PTAM (GPU (no events) vs. CPU (event-aided)) qualitative analysis. All trajectories are transformed to a reference frame as the ground truth poses using the extrinsic parameters, followed by an alignment with all poses by Umeyama’s SE(3) method implemented in \cite{grupp2017evo}. Large-scale trajectories show high-quality loop closure detection in the case of R2D2 on GPU. The small-scale trajectory shows the high accuracy of the event-aided version of DH-PTAM.}
\label{fig:qual-rpe}
\end{figure*}

We perform a thorough, comprehensive evaluation during navigation in real-world, large-scale, and small-scale areas in challenging settings. In subsection \ref{large}, we compare DH-PTAM with other RGB image-based and event-based/-aided methods on the HDR large-scale sequences of the publicly available dataset VECtor \cite{gao2022vector} due to its high-quality ground truth values and sensors calibration parameters. In subsection \ref{small}, we evaluate the small-scale (mocap-) sequences of TUM-VIE \cite{klenk2021tum} to test the quality of the DH-PTAM spatio-temporal synchronization method with degraded event camera calibration parameters. Moreover, the first 45 frames of TUM-VIE sequences suffer a high over-/under-exposure global shutter alternation, which tests the DH-PTAM's pose estimation stability. We perform a comparative quantitative analysis to evaluate the accuracy of our system in Table \ref{tab:eval} and a qualitative analysis in Fig. \ref{fig:qual-rpe}. The accuracy of DH-PTAM is measured with absolute trajectory error (ATE), and relative pose error (RPE) metrics calculated using the baseline SLAM evaluation tool \cite{grupp2017evo}.

\begin{table*}
\begin{center}
\caption{DH-PTAM Quantitative Comparison Against the best performing open-source State-of-the-art SLAM Systems using ATE [m] metric. The upper sub-table is for Standard Stereo VIO Methods, the middle is for event-based VO/VIO Methods, and the lower is for DH-PTAM with RPE [m] metric. \textbf{Bold} denotes best performing, \underline{Underline} for second best performing, and ($\times$) denotes failure}
\label{tab:eval}
\resizebox{\linewidth}{!}{%
\begin{tabular}{| c | cccccc | ccccc | cc |}
\hline
\multirow{3}{*}{Method} & \multicolumn{6}{c|}{VECtor sequences \cite{gao2022vector}} & \multicolumn{5}{c|}{TUM-VIE sequences \cite{klenk2021tum}} & {Mean} & {Mean}\\
 & corridors & corridors & units & units & school & school  & mocap & mocap & mocap & mocap & mocap & {VECtor} & {TUM-VIE} \\
  & dolly & walk & dolly & scooter & dolly & scooter  & 1d-trans & 3d-trans & 6dof & desk & desk2 & {large-scale} & {small-scale} \\
\hline
ORB-SLAM3 (SVIO) \cite{campos2021orb} & \textbf{0.802} & \underline{1.031} & 18.063 & 14.504 & \textbf{0.921} & \underline{0.752} & \underline{0.007} & 0.012 & 0.018 & \textbf{0.007} & 0.025 & 6.012 & \underline{0.013} \\ 
BASALT (SVIO) \cite{usenko2019visual} & 1.625 & 2.152 & 11.151 & 13.256 & 1.852 & 1.482 & \textbf{0.003} & 0.009 & \textbf{0.014} & 0.016 & \underline{0.011} & 5.253 & \textbf{0.011} \\ 
VINS-Fusion (SVIO) \cite{qin2019general} & \underline{1.464} & \textbf{0.392} & 10.391 & 11.471 & 1.791 & \textbf{0.562} & 0.011 & 0.011 & \underline{0.017} & 0.058 & 0.013 & 4.345 & 0.022 \\ 
\hline
EVO (Mono Events) \cite{rebecq2016evo} & $\times$ & $\times$ & $\times$ & $\times$ & $\times$ & $\times$ & 0.075 & 0.125 & 0.855 & 0.541 & 0.752 & $\times$ & 0.470 \\ 
ESVO (Stereo Events) \cite{zhou2021event} & $\times$ & $\times$ & $\times$ & $\times$ & 13.710 & 9.830 & 0.009 & 0.028 & 0.058 & 0.033 & 0.032 & 11.77 & 0.032 \\ 
Ultimate SLAM (EVIO)$^+$ \cite{vidal2018ultimate} & $\times$ & $\times$ & $\times$ & $\times$ & $\times$ & 6.830 & 0.039 & 0.047 & 0.353 & 0.195 & 0.341 & 6.830 & 0.195 \\ 
\hline 
\textbf{DH-PTAM (Stereo Fusion)} & 1.884 & 1.299 & \textbf{5.274} & \underline{8.433} & \underline{1.093} & 0.796 & 0.103 & \underline{0.007} & 0.024 & 0.016 & 0.015 & \underline{3.130} & 0.033 \\
(SuperPoint on CPU) - RPE ($\sigma$) & 0.073 & 0.038 & 0.055 & 0.149 & 0.178 & 0.074 & 0.006 & 0.007 & 0.009 & 0.009 & 0.007 & 0.095 & 0.008 \\
\textbf{DH-PTAM$^*$  (Stereo Image)} & 1.841 & 1.543 & \underline{5.738} & \textbf{5.010} & 1.559 & 0.877 & 0.099 & \textbf{0.004} & 0.045 & \underline{0.011} & \textbf{0.008} & \textbf{2.761} & 0.033 \\
(R2D2 on GPU) - RPE ($\sigma$) &  0.116 & 0.141 & 0.134 & 0.308 & 0.202 & 0.331 & 0.014 & 0.020 & 0.022 & 0.023 & 0.021 & 0.205 & 0.020 \\
\hline 
\end{tabular}%
}
\end{center}
\footnotesize{$^+$ IMU sensor is included since it is integrated into the front-end and cannot be separated for a fair comparison with EVO, ESVO, and DH-PTAM (ours).\\$^*$ in this ablation case study, the SuperPoint detector is replaced with the R2D2 detector (trained for SLAM tasks), leveraging the GPU performance.}
\end{table*}

\begin{figure*}[t]
    \centering
    \resizebox{\linewidth}{!}{
    \begin{tabular}{cccccc}
        & {EVO \cite{rebecq2016evo}} & {EDS \cite{hidalgo2022event}} & {DH-PTAM (Ours)} & {EDS (point cloud) \cite{hidalgo2022event}} & {DH-PTAM (point cloud)} \\

        \multirow{-5.5}{*}{\rotatebox[origin=c]{90}{{monitor}}} &
        \includegraphics[width=0.19\textwidth]{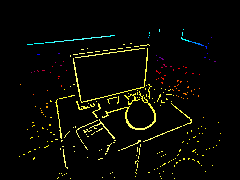} &
        \includegraphics[width=0.19\textwidth]{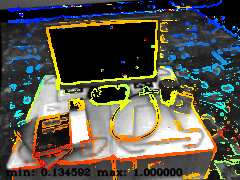} & 
        \includegraphics[width=0.19\textwidth]{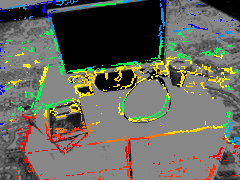} &
        \includegraphics[width=0.19\textwidth]{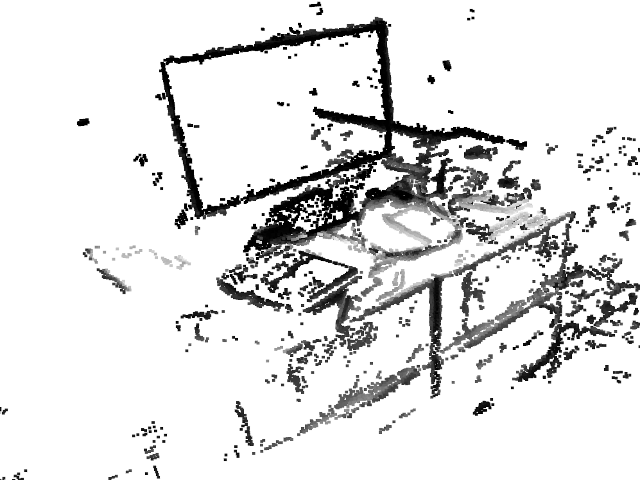} &
        \includegraphics[width=0.19\textwidth]{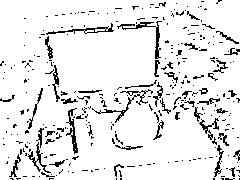} \\
         
        \multirow{-5.5}{*}{\rotatebox[origin=c]{90}{{boxes}}} &
        \includegraphics[width=0.19\textwidth]{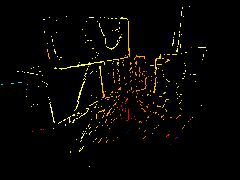} &
        \includegraphics[width=0.19\textwidth]{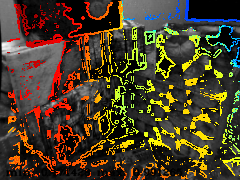} & 
        \includegraphics[width=0.19\textwidth]{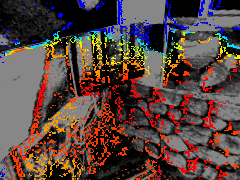} &
        \includegraphics[width=0.19\textwidth]{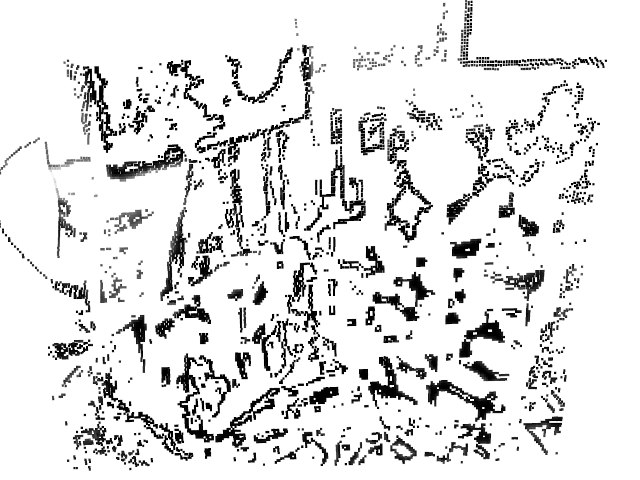} &
        \includegraphics[width=0.19\textwidth]{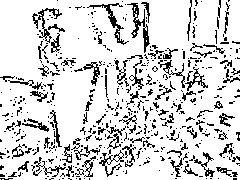} \\
        
        \multirow{-5.5}{*}{\rotatebox[origin=c]{90}{{desk}}} &
        \includegraphics[width=0.19\textwidth]{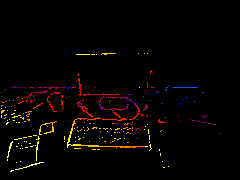} &
        \includegraphics[width=0.19\textwidth]{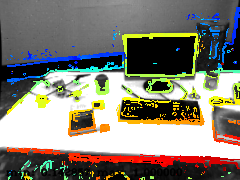} & 
        \includegraphics[width=0.19\textwidth]{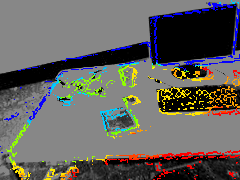} &
        \includegraphics[width=0.19\textwidth]{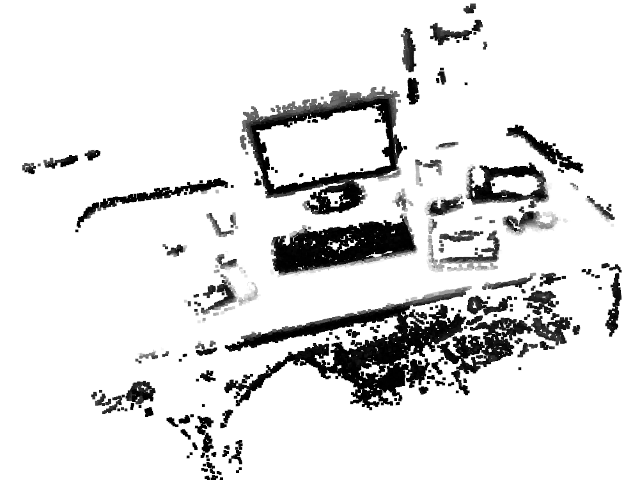} &
        \includegraphics[width=0.19\textwidth]{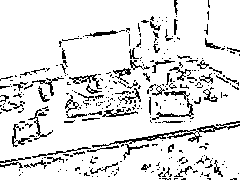} \\

        \multirow{-5.5}{*}{\rotatebox[origin=c]{90}{{bin}}} &
        \includegraphics[width=0.19\textwidth]{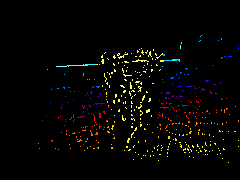} &
        \includegraphics[width=0.19\textwidth]{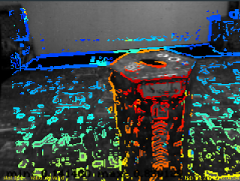} & 
        \includegraphics[width=0.19\textwidth]{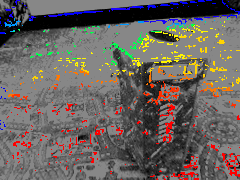} &
        \includegraphics[width=0.19\textwidth]{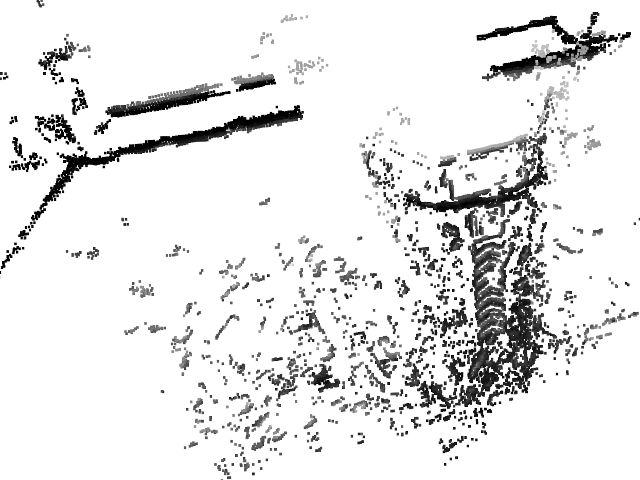} &
        \includegraphics[width=0.19\textwidth]{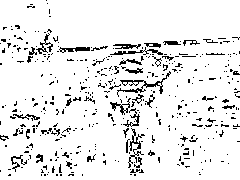}
        
    \end{tabular}%
    }
    \caption{Stereo DAVIS sequences evaluation from \cite{zhou2018semi} with $\delta{P^h_{align}}=[0,0]^\top$ since both sensors (event-based and frame-based) share the same photoreceptor pixel array. Initial three columns present pseudo-colored inverse depth maps: EVO (yellow-near, blue-far), EDS and DH-PTAM (red-near, blue-far). All sequences span a 1-7 m depth range. 3D point clouds from EDS and DH-PTAM, utilizing grayscale keyframe values, are presented in the last two columns.}
    \label{fig:qual-pcd}
\end{figure*}

To prevail the advantages of complementing the sensor stack with events information, we compare our event-aided stereo visual odometry solution (DH-PTAM) to the latest best-performing open-source visual-inertial systems in literature in Table \ref{tab:eval}. Table \ref{tab:param} gives the system parameters configuration for large-scale and small-scale sequences. We keep these parameters constant for all sequences of the same scale group without an online fine-tuning process.

\begin{table}[!t]
\caption{DH-PTAM Parameters Configuration \label{tab:param}}
\begin{center}
\resizebox{\linewidth}{!}{
\begin{tabular}{lcc}
\hline
Parameter & VECtor sequences & TUM-VIE sequences\\
\hline
$\delta{P^h_{align}}$ - Left &  (-160, -235) [px] & (355, 40) [px] \\
$\delta{P^h_{align}}$ - Right &  (-160, -235) [px] & (375, 45) [px] \\
frustum\_near & 0.1 [m] & 0.1 [m] \\
frustum\_far & 30.0 [m] & 5.0 [m] \\
matching\_cell\_size & 15 [px] & 15 [px] \\
matching\_neighborhood & 2 [px] & 1.8 [px] \\
matching\_distance & 25 [px] & 15 [px] \\
\hline
\end{tabular}
}
\end{center}
\end{table}

All experiments are performed on the CPU and the GPU of a 16 GB RAM laptop computer running 64-bit Ubuntu 20.04.3 LTS with AMD(R) Ryzen 7 4800h ×16 cores 2.9 GHz processor and a Radeon RTX NV166 Renoir graphics card. Table \ref{tab:cmplx} reports a detailed computational complexity analysis for our DH-PTAM system with minimal and maximal system requirements. The high CPU load observed when detecting SuperPoint and R2D2 features can be attributed to the algorithms' design, which prioritizes feature quality and robustness over computational efficiency. This trade-off is often necessary for computer vision research, where high-quality results are crucial for many applications but come at the cost of increased computational complexity. The back-end runs with real-time performance, and it is recommended to run the front-end on a GPU to achieve a memory efficient, faster, and more stable performance.

\begin{table}[!t]
\caption{Computational Complexity Analysis on CPU vs. GPU \label{tab:cmplx}}
\begin{center}
\resizebox{\linewidth}{!}{
\begin{tabular}{lllll}
\hline
Thread & \#Tasks & Operation & CPU [ms] & GPU [ms]\\
\hline
Front-end & 3 & Stereo E3CT Construction & 25-30 & 76-172 \\
     &   & Events-Frames Fusion & 439-521 & 191-352\\
     & ($\approx$2$\times$10K)  & SuperPoint Detection & 2478-3256 & 521-1752\\
      & ($\approx$2$\times$4K)  & R2D2 Detection & 8532-8752 & 1067-2254\\
Bootstrapping & 1 & Initialize the Map & 106-143 & 53-120\\
Tracking & 2 & Spatio-temporal Matching & 161-215 & 142-172\\
         & (10 iter.)  & Pose Refinement & 11-15 & 10-12\\
Mapping & 2 & Update Map & 1-3 & 0.452-1\\
        & (30 iter.) & Local Bundle-Adjustment & 1-4 & 1-2\\
Loop-closing & 3 & Loop Detection & 14-20 & 10-15\\
 &  & Compute and Validate & 2-5 & 1-3\\
 & (30 iter.) & Pose Graph Optimization & 1-3 & 0.524-2\\
\hline
 End-to-End & 11 & SuperPoint Detector & 3153-4208 & 356-1962\\
  &  & R2D2 Detector & 8560-9125 & 1226-2486\\
\hline
\end{tabular}
}
\end{center}
\end{table}

\subsection{VECtor large-scale experiments \label{large}}

We notice a prominent estimation failure in Table \ref{tab:eval} while evaluating the event-based methods EVO, ESVO and Ultimate SLAM on the large-scale sequences. Numerous factors may contribute to the failure of these systems, including stringent initialization requirements. For instance, the system EVO necessitates running in a sensor-planar scene for several seconds to bootstrap the system. Additionally, these systems are susceptible to parameter tuning, as demonstrated by using different parameters for different sequences in the same scenarios, even within their open-source projects.

Table \ref{tab:eval} shows a good performance for DH-PTAM compared to the competing VI-SLAM systems. Although Fig. \ref{fig:qual-rpe} shows high visual drifts for our vision-only system in the case of units sequences, DH-PTAM could outperform the VI-SLAM systems based on the ATE metric. Fig. \ref{fig:qual-rpe} gives an overview of the high-quality loop detection of DH-PTAM in the case of corridors sequences. Loop detection failure can be noticed only when the RAM overflows while running the system with enormous point clouds, as in the case of units sequences. We provide trajectory smoothing and post-processing script with our open-source implementation to join estimated trajectory increments in case of RAM overflow failures. 

\subsection{TUM-VIE small-scale experiments \label{small}}

As noticed in \cite{multieventcam}, the calibrationA (mocap-desk, mocap-desk2) sequences have more accurate depth estimation results than calibrationB (rest of mocap and TUM-VIE large-scale) sequences due to the significant calibration errors in the latter. Hence, we perform our comparative evaluation on TUM-VIE small-scale (mocap-) sequences using calibrationA parameters. Although the same high-quality calibrationA parameters apply to both desk2 and desk sequences with the same spiral motion, DH-PTAM performs the best with desk2 sequence but the worst with desk sequence. This occurs since the scene of the desk sequence is bounded by a close-by white wall that strict the depth, and hence DH-PTAM front-end detects low quality and fewer features for desk than desk2. Table \ref{tab:eval} shows that the more DoF excited (6dof, desk2) and the consistent loops detection (1d-trans), the better the pose estimation quality.

\subsection{Ablation experiments}

\textbf{No event streams ($\beta=0$)}. In Table \ref{tab:eval}, we show an ablation study where we run DH-PTAM on stereo images. We notice estimation failure with all the conventional and learning-based feature detectors except R2D2. Although the ATE metric shows slightly better results without using events, the RPE metric shows much more accurate values when using events. These better ATE values are due to the high performance of the GPU in loop-closures using R2D2 features (see Fig. \ref{fig:qual-rpe}).

\textbf{Front-end depth maps}. Qualitatively in Fig. \ref{fig:qual-pcd}, we rigorously assess the front-end of our events-frames fusion technique by comparing its inverse depth maps against established methods such as EVO and EDS. By channeling the stereo fusion frames through the SGM matching method \cite{4359315}, our method capitalizes on direct front-end fusion where the inverse depth maps are derived from events projected onto the RGB frames. Remarkably, the point clouds generated through our method exhibited richer details compared to those of EDS. This affirms the potential of our method to set a new benchmark in event-frame fusion paradigms.

\section{Conclusion\label{sec:concl}}

In this paper, we presented the DH-PTAM system for robust parallel tracking and mapping in dynamic environments using stereo images and event streams. The proposed system builds upon the principles of S-PTAM and extends it with a learning-based approach to handle the sparse and noisy nature of event-based sensors while leveraging the rich information provided by fusion frames. Our experiments demonstrate that DH-PTAM outperforms state-of-the-art visual-SLAM methods, particularly in challenging scenarios such as HDR and occlusions. The proposed system can achieve better performance on a GPU and provides a scalable and accurate solution for 3D reconstruction and pose estimation. Future work includes investigating the potential of improving the spatial synchronization through online optimization and exploring the integration of an adaptive paradigm for choosing the temporal window width to further boost robustness and system latency. DH-PTAM proves to have the potential to provide robust and accurate 3D mapping and localization, which are crucial for the successful operation of long-term navigation systems.

\bibliographystyle{IEEEtran}
\bibliography{RefList}

\begin{thebibliography}{10}
\providecommand{\url}[1]{#1}
\csname url@rmstyle\endcsname
\providecommand{\newblock}{\relax}
\providecommand{\bibinfo}[2]{#2}
\providecommand\BIBentrySTDinterwordspacing{\spaceskip=0pt\relax}
\providecommand\BIBentryALTinterwordstretchfactor{4}
\providecommand\BIBentryALTinterwordspacing{\spaceskip=\fontdimen2\font plus
\BIBentryALTinterwordstretchfactor\fontdimen3\font minus
  \fontdimen4\font\relax}
\providecommand\BIBforeignlanguage[2]{{%
\expandafter\ifx\csname l@#1\endcsname\relax
\typeout{** WARNING: IEEEtran.bst: No hyphenation pattern has been}%
\typeout{** loaded for the language `#1'. Using the pattern for}%
\typeout{** the default language instead.}%
\else
\language=\csname l@#1\endcsname
\fi
#2}}

\bibitem{4538852}
G.~Klein and D.~Murray, ``Parallel tracking and mapping for small ar
  workspaces,'' in \emph{2007 6th IEEE and ACM International Symposium on Mixed
  and Augmented Reality}, 2007, pp. 225--234.

\bibitem{pire2017s}
T.~Pire, T.~Fischer, G.~Castro, P.~De~Crist{\'o}foris, J.~Civera, and J.~J.
  Berlles, ``S-ptam: Stereo parallel tracking and mapping,'' \emph{Robotics and
  Autonomous Systems}, vol.~93, pp. 27--42, 2017.

\bibitem{detone2018superpoint}
D.~DeTone, T.~Malisiewicz, and A.~Rabinovich, ``Superpoint: Self-supervised
  interest point detection and description,'' in \emph{Proceedings of the IEEE
  conference on computer vision and pattern recognition workshops}, 2018, pp.
  224--236.

\bibitem{revaud2019r2d2}
J.~Revaud, C.~De~Souza, M.~Humenberger, and P.~Weinzaepfel, ``R2d2: Reliable
  and repeatable detector and descriptor,'' \emph{Advances in neural
  information processing systems}, vol.~32, 2019.

\bibitem{teed2021droid}
Z.~Teed and J.~Deng, ``Droid-slam: Deep visual slam for monocular, stereo, and
  rgb-d cameras,'' \emph{Advances in Neural Information Processing Systems},
  vol.~34, pp. 16\,558--16\,569, 2021.

\bibitem{zhang2017active}
Z.~Zhang, C.~Forster, and D.~Scaramuzza, ``Active exposure control for robust
  visual odometry in hdr environments,'' in \emph{2017 IEEE International
  Conference on Robotics and Automation (ICRA)}.\hskip 1em plus 0.5em minus
  0.4em\relax IEEE, 2017, pp. 3894--3901.

\bibitem{hidalgo2022event}
J.~Hidalgo-Carri{\'o}, G.~Gallego, and D.~Scaramuzza, ``Event-aided direct
  sparse odometry,'' in \emph{Proceedings of the IEEE/CVF Conference on
  Computer Vision and Pattern Recognition}, 2022, pp. 5781--5790.

\bibitem{sun2021autonomous}
S.~Sun, G.~Cioffi, C.~De~Visser, and D.~Scaramuzza, ``Autonomous quadrotor
  flight despite rotor failure with onboard vision sensors: Frames vs.
  events,'' \emph{IEEE Robotics and Automation Letters}, vol.~6, no.~2, pp.
  580--587, 2021.

\bibitem{gao2022vector}
L.~Gao, Y.~Liang, J.~Yang, S.~Wu, C.~Wang, J.~Chen, and L.~Kneip, ``{VECtor}: A
  versatile event-centric benchmark for multi-sensor slam,'' \emph{IEEE
  Robotics and Automation Letters}, vol.~7, no.~3, pp. 8217--8224, 2022.

\bibitem{klenk2021tum}
S.~Klenk, J.~Chui, N.~Demmel, and D.~Cremers, ``Tum-vie: The tum stereo
  visual-inertial event dataset,'' in \emph{2021 IEEE/RSJ International
  Conference on Intelligent Robots and Systems (IROS)}.\hskip 1em plus 0.5em
  minus 0.4em\relax IEEE, 2021, pp. 8601--8608.

\bibitem{ibiscape}
A.~Soliman, F.~Bonardi, D.~Sidib{\'e}, and S.~Bouchafa, ``{IBISCape}: A
  simulated benchmark for multi-modal {SLAM} systems evaluation in large-scale
  dynamic environments,'' \emph{Journal of Intelligent {\&} Robotic Systems},
  vol. 106, no.~3, p.~53, Oct 2022.

\bibitem{rebecq2016evo}
H.~Rebecq, T.~Horstsch{\"a}fer, G.~Gallego, and D.~Scaramuzza, ``Evo: A
  geometric approach to event-based 6-dof parallel tracking and mapping in real
  time,'' \emph{IEEE Robotics and Automation Letters}, vol.~2, no.~2, pp.
  593--600, 2016.

\bibitem{sironi2018hats}
A.~Sironi, M.~Brambilla, N.~Bourdis, X.~Lagorce, and R.~Benosman, ``Hats:
  Histograms of averaged time surfaces for robust event-based object
  classification,'' in \emph{Proceedings of the IEEE Conference on Computer
  Vision and Pattern Recognition}, 2018, pp. 1731--1740.

\bibitem{gehrig2019end}
D.~Gehrig, A.~Loquercio, K.~G. Derpanis, and D.~Scaramuzza, ``End-to-end
  learning of representations for asynchronous event-based data,'' in
  \emph{Proceedings of the IEEE/CVF International Conference on Computer
  Vision}, 2019, pp. 5633--5643.

\bibitem{gehrig2020eklt}
D.~Gehrig, H.~Rebecq, G.~Gallego, and D.~Scaramuzza, ``Eklt: Asynchronous
  photometric feature tracking using events and frames,'' \emph{International
  Journal of Computer Vision}, vol. 128, no.~3, pp. 601--618, 2020.

\bibitem{vidal2018ultimate}
A.~R. Vidal, H.~Rebecq, T.~Horstschaefer, and D.~Scaramuzza, ``Ultimate slam?
  combining events, images, and imu for robust visual slam in hdr and
  high-speed scenarios,'' \emph{IEEE Robotics and Automation Letters}, vol.~3,
  no.~2, pp. 994--1001, 2018.

\bibitem{kim2016real}
H.~Kim, S.~Leutenegger, and A.~J. Davison, ``Real-time 3d reconstruction and
  6-dof tracking with an event camera,'' in \emph{European conference on
  computer vision}.\hskip 1em plus 0.5em minus 0.4em\relax Springer, 2016, pp.
  349--364.

\bibitem{cadena2021spade}
P.~R.~G. Cadena, Y.~Qian, C.~Wang, and M.~Yang, ``Spade-e2vid:
  Spatially-adaptive denormalization for event-based video reconstruction,''
  \emph{IEEE Transactions on Image Processing}, vol.~30, pp. 2488--2500, 2021.

\bibitem{campos2021orb}
C.~Campos, R.~Elvira, J.~J.~G. Rodr{\'\i}guez, J.~M. Montiel, and J.~D.
  Tard{\'o}s, ``{OrbSLAM3}: An accurate open-source library for visual,
  visual--inertial, and multimap slam,'' \emph{IEEE Transactions on Robotics},
  vol.~37, no.~6, pp. 1874--1890, 2021.

\bibitem{koniusz2013comparison}
P.~Koniusz, F.~Yan, and K.~Mikolajczyk, ``Comparison of mid-level feature
  coding approaches and pooling strategies in visual concept detection,''
  \emph{Computer vision and image understanding}, vol. 117, no.~5, pp.
  479--492, 2013.

\bibitem{GalvezTRO12}
D.~G\'alvez-L\'opez and J.~D. Tard\'os, ``Bags of binary words for fast place
  recognition in image sequences,'' \emph{IEEE Transactions on Robotics},
  vol.~28, no.~5, pp. 1188--1197, October 2012.

\bibitem{app10010140}
Z.~Ji, F.~Wang, X.~Gao, L.~Xu, and X.~Hu, ``Ssnet: Learning mid-level image
  representation using salient superpixel network,'' \emph{Applied Sciences},
  vol.~10, no.~1, 2020.

\bibitem{zhou2021event}
Y.~Zhou, G.~Gallego, and S.~Shen, ``Event-based stereo visual odometry,''
  \emph{IEEE Transactions on Robotics}, vol.~37, no.~5, pp. 1433--1450, 2021.

\bibitem{kueng2016low}
B.~Kueng, E.~Mueggler, G.~Gallego, and D.~Scaramuzza, ``Low-latency visual
  odometry using event-based feature tracks,'' in \emph{2016 IEEE/RSJ
  International Conference on Intelligent Robots and Systems (IROS)}.\hskip 1em
  plus 0.5em minus 0.4em\relax IEEE, 2016, pp. 16--23.

\bibitem{9710962}
L.~Wang, Y.~Wang, L.~Wang, Y.~Zhan, Y.~Wang, and H.~Lu, ``Can scale-consistent
  monocular depth be learned in a self-supervised scale-invariant manner?'' in
  \emph{2021 IEEE/CVF International Conference on Computer Vision (ICCV)},
  2021, pp. 12\,707--12\,716.

\bibitem{grupp2017evo}
M.~Grupp, ``evo: Python package for the evaluation of odometry and slam,''
  Note: \url{https://github.com/MichaelGrupp/evo}, 2017.

\bibitem{usenko2019visual}
V.~Usenko, N.~Demmel, D.~Schubert, J.~St{\"u}ckler, and D.~Cremers,
  ``Visual-inertial mapping with non-linear factor recovery,'' \emph{IEEE
  Robotics and Automation Letters}, vol.~5, no.~2, pp. 422--429, 2019.

\bibitem{qin2019general}
T.~Qin, J.~Pan, S.~Cao, and S.~Shen, ``A general optimization-based framework
  for local odometry estimation with multiple sensors,'' \emph{arXiv preprint
  arXiv:1901.03638}, 2019.

\bibitem{zhou2018semi}
Y.~Zhou, G.~Gallego, H.~Rebecq, L.~Kneip, H.~Li, and D.~Scaramuzza,
  ``Semi-dense 3d reconstruction with a stereo event camera,'' in
  \emph{Proceedings of the European conference on computer vision (ECCV)},
  2018, pp. 235--251.

\bibitem{multieventcam}
S.~Ghosh and G.~Gallego, ``Multi-event-camera depth estimation and outlier
  rejection by refocused events fusion,'' \emph{Advanced Intelligent Systems},
  vol.~4, no.~12, p. 2200221, 2022.

\bibitem{4359315}
H.~Hirschmuller, ``Stereo processing by semiglobal matching and mutual
  information,'' \emph{IEEE Transactions on Pattern Analysis and Machine
  Intelligence}, vol.~30, no.~2, pp. 328--341, 2008.

\end{thebibliography}

\vfill

\end{document}